\documentclass[letterpaper, 10 pt, conference]{ieeeconf}
\IEEEoverridecommandlockouts
\usepackage{cite}
\usepackage{amsmath,amssymb,amsfonts}
\usepackage{algorithmic}
\usepackage{graphicx}
\usepackage{textcomp}
\usepackage{xcolor}
\def\BibTeX{{\rm B\kern-.05em{\sc i\kern-.025em b}\kern-.08em
    T\kern-.1667em\lower.7ex\hbox{E}\kern-.125emX}}
\begin{document}

\title{\LARGE \bf
FlipWalker: Jacob's Ladder toy-inspired robot for locomotion across diverse, complex terrain
}

\author{Diancheng Li$^{1,2}$, Nia Ralston$^{1,2}$, Bastiaan Hagen$^{1,2}$, Phoebe Tan$^{1}$ and Matthew A. Robertson$^{*1,2}$
\thanks{$^{1}$Department of Mechanical and Materials Engineering, Queen's University, K7L 3N6, Canada.}%
\thanks{$^{2}$Ingenuity Labs Research Institute, Queen's University, K7L 2N9, Canada.}
\thanks{Corresponding author e-mail: {\tt\small m.robertson@queensu.ca}.}        
}

\maketitle

\begin{abstract}
This paper introduces FlipWalker, a novel underactuated robot locomotion system inspired by Jacob's Ladder illusion toy, designed to traverse challenging terrains where wheeled robots often struggle. Like the Jacob’s Ladder toy, FlipWalker features two interconnected segments joined by flexible cables, enabling it to pivot and flip around singularities in a manner reminiscent of the toy’s cascading motion. Actuation is provided by motor-driven legs within each segment that push off either the ground or the opposing segment, depending on the robot’s current configuration. A physics-based model of the underactuated flipping dynamics is formulated to elucidate the critical design parameters governing forward motion and obstacle clearance or climbing. The untethered prototype weighs 0.78 kg, achieves a maximum flipping speed of 0.2 body lengths per second. Experimental trials on artificial grass, river rocks, and snow demonstrate that FlipWalker’s flipping strategy, which relies on ground reaction forces applied normal to the surface, offers a promising alternative to traditional locomotion for navigating irregular outdoor terrain.
\end{abstract}

\section{Introduction}
In the realm of robotic locomotion, an evolving array of techniques has been harnessed to navigate increasingly complex terrains. From conventional wheel systems \cite{chung_wheeled_2016,zheng2022mathbf} and track systems \cite{lee_series_2016,kalra_wall_2006} to bio-inspired strategies such as climbing \cite{chen_soft_2023,liao_soft_2020,uno_hubrobo_2021}, walking \cite{guan_design_2022,todd_walking_2013,torres-pardo_legged_2022,fong_mechanical_2005,paluska_design_nodate,guizzo_by_2019}, hopping \cite{yim_hopping_nodate,fiorini_development_2003,hyon_development_2002,brown_bow_1998,yoshimitsu_micro-hopping_2003} rolling \cite{armour_rolling_2006,liang2020freebot,gim2024ringbot} and also flipping\cite{seo2013flipbot,chen_soft_2023} , a diverse range of mobility solutions tailored to specific environmental challenges have been explored. In response to ongoing efforts to understand and mitigate climate change, its effect on the natural world and on civilization
the demand for adaptable and versatile robotic platforms in various applications, ranging from biodiversity surveying \cite{maslin_underwater_2021,aucone_drone_assisted_2023,angelini_robotic_2023} to disaster response \cite{bogue_disaster_2019,nagatani_emergency_2013}, has intensified. In light of these developments, the importance of innovative locomotion concepts becomes even more pronounced. 

Recently, the field of robotics has begun to draw inspiration from various diverse sources, including nature and traditional art forms. For instance, the art of origami, a traditional Japanese paper-folding technique, and kirigami, another Japanese art form that involves paper cutting, have been significant influences in the design of soft robotic grippers and other robotic structures \cite{yang_grasping_2021, rus_spotlight_2018, robertson2021soft, li2023origami}. Prior to these advances, the work in curved-folding mechanisms demonstrates how the principles of compliant rolling and deployable geometries can further expand the functional capabilities of origami and kirigami-inspired systems\cite{nelson_curved-folding-inspired_2016, halverson2010tension}.

This paper introduces a novel approach to robotic locomotion, drawing inspiration from an unexpected source: the mechanics of the Jacob's Ladder (JL) toy \cite{noauthor_scientific_1889, noauthor_physics_nodate}. Unlike conventional methods, this new approach, referred to as FlipWalker, capitalizes on the intriguing dynamics of the Jacob's Ladder to achieve forward locomotion through sequential body flipping. By unveiling the latent potential within the mechanics of this centuries-old toy, the study not only pushes the boundaries of robotic mobility but also introduces a distinctive and promising strategy for traversing challenging landscapes.
The Jacob’s Ladder toy is characterized by rigid body segments linked together by flexible bands, creating a unique structure that seamlessly transitions its rotational axes around a mechanical singularity. Historically recognized for its ability to engage the curiosity of children and confound adults across centuries, there are even accounts suggesting the discovery of such a toy in King Tut's tomb, and its application has been primarily relegated to amusement rather than locomotion \cite{immel_jacobs_2019,noauthor_david_nodate}.

\begin{figure*}[htbp]
  \centering
  \includegraphics[width=\textwidth]{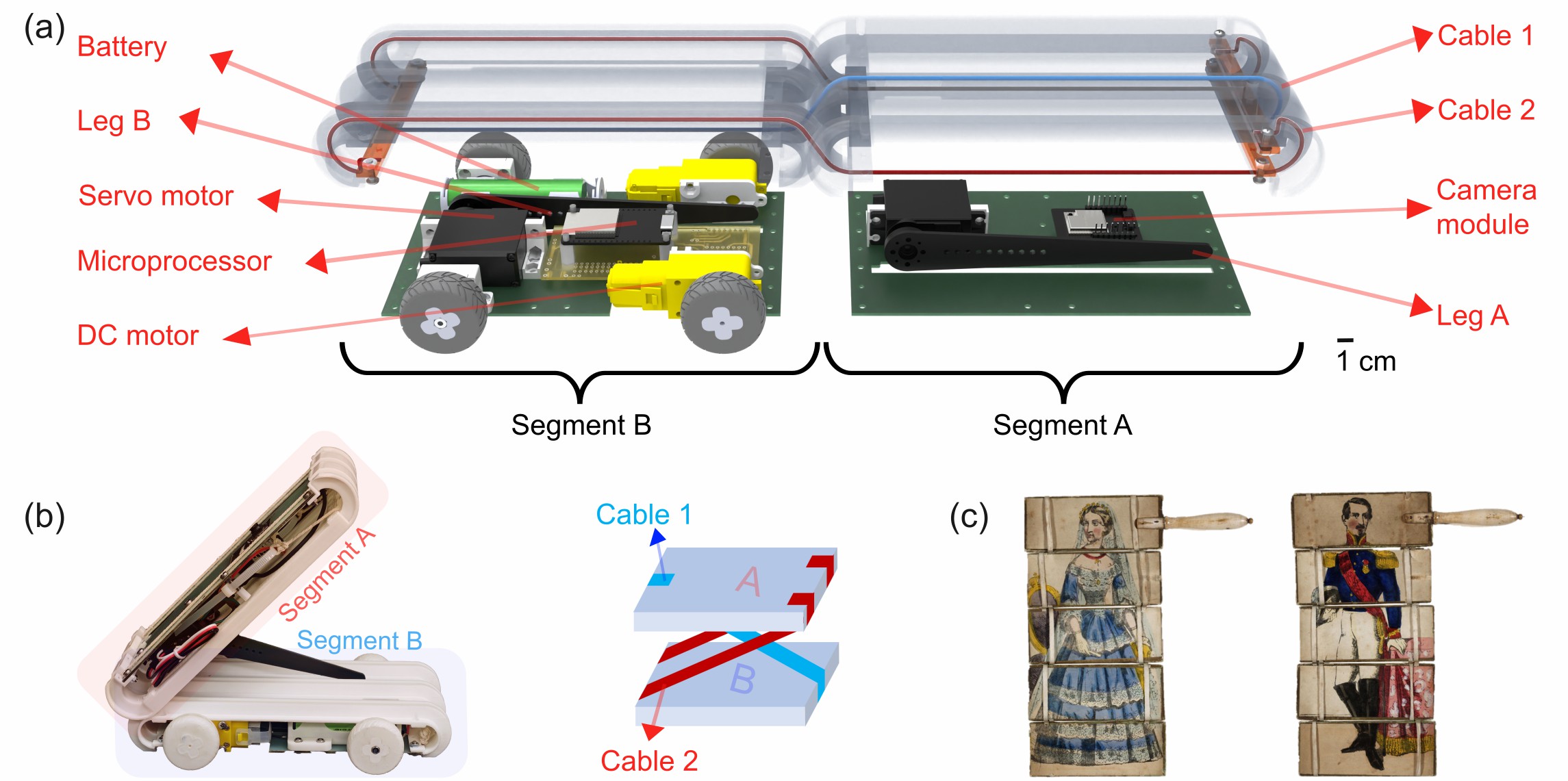}
  \caption{The FlipWalker concept borrows its mechanism from the Jacob's Ladder toy: rigid body segments linked together by flexible cables which enable seamless switching of rotational axes through the transition across a mechanical singularity. (a) exploded view of the FlipWalker prototype robot CAD model, (b) robot configuration, (c) Jacob’s ladder of Queen Victoria and Prince Albert, 1840, \textit{Toronto Public Library Special Collections.}}
  \label{fig:main}
\end{figure*}

The Jacob's Ladder concept dates back to at least 1515 AD, with the earliest known reference found in Luca Pacioli's unpublished manuscript "De viribus quantitatis." The mechanism also appears in the period painting "Boy with Puzzle" by Bernardino Luini \cite{hirth_luca_2015}. Traditionally made from wooden blocks connected by flat ribbons, the toy operates by inverting the top block, triggering a cascading flipping motion down the chain. Although the motion creates the illusion that a block travels down the chain, each block only reverses its orientation in place.

Unlike existing flipping mechanisms where robots momentarily lose ground contact during the flipping motion \cite{geng_dynamics_2005}, the FlipWalker ensures that at least one segment of the robot maintains contact with the ground throughout the flipping process. Moreover, as the FlipWalker encompasses additional segments, the sequential flipping ensures that only one segment is in the flipping state at any given time, guaranteeing that a significant portion of the robot always stays grounded. This design choice not only enhances stability but also allows for continuous interaction with the terrain, enabling improved control and adaptability.

The Toy derives its behavior from its unique construction which forms a kinematic singularity when two blocks are parallel (with one directly above the other). The connecting cables, wound around the blocks in alternating directions, act as both a flexure hinge and a mechanical constraint, allowing adjacent blocks to rotate relative to each other about a center of rotation at their ends. On opposite sides of the singularity, the axis of rotation between blocks can therefore seamlessly switch from one end to the other. The switching of center of rotation occurs passively and without the use of any traditional type of clutching mechanism - rather it simply results from the geometry and kinematics of the flexible cable linkage. 

In this work, we present a novel robotic locomotion platform—FlipWalker—that employs a unique flipping mechanism inspired by the Jacob’s Ladder toy to achieve both locomotion and obstacle traversal. The key contributions of this work are summarized as follows:

\begin{itemize}
    \item We develop a unique robotic locomotion system that enables both linear locomotion and climbing over obstacles (or even other robots) by cyclically flipping interconnected body segments.
    \item We demonstrate mobility across diverse, complex terrains using a flipping-based strategy that relies on normal-force ground push-off, distinct from the tangential tractive forces typically exploited by wheeled or tracked robots.
    \item We formulate a kinematic and geometric model that relates key design parameters to obstacle clearance ability and flipping dynamics, offering design insights for future iterations of this locomotion platform.
\end{itemize}

This work also highlights a fundamental distinction between the Jacob’s Ladder toy and FlipWalker’s locomotion mechanism. In the original toy, the cascading motion is a visual illusion where the blocks themselves do not actually change order. In contrast, FlipWalker’s two body segments physically exchange positions with each flip, generating true forward displacement. This cyclic flipping and position exchange effect serves as the core principle enabling continuous locomotion along the ground.

\section{Design}

\subsection{Locomotion concept robot prototype}

An untethered, remotely activated robot prototype was first developed to demonstrate the locomotion principle. The robot consists of two rigid body segments attached only by a set of flexible, inelastic cables (Fig.~\ref{fig:main}). The primary structure of each body segment of the robot is constructed from 3D printed shell, with a laser-cut base to mount all components. The ends of each of the cables are fixed to both body segments (one end per segment) and alternately wrapped around the segments to produce the kinematic singularity to which the robot owes its unique ability to switch axes of rotation. Although there are three cables used in total, effectively, there are only two cable configurations, as one cable is split into two smaller cables that follow the same path wound around the body segments. Notably, cable 1 (blue one in Fig.~\ref{fig:main}) also plays a role in the communication and power supply between the two segments.

Within each body segment, there are two single DC gearmotors (Robotshop yellow motors) and one servo motor (HOOYIJ DS3225). The servo motor is located within each segment close to the edge, with a leg attached to each side of the servo leg. The legs are 3D printed from carbon-fiber filament, and each rotates about the motor axis, which is parallel to the axis of rotation formed by the cable constraints between segments. The symmetric placement of the servo motor and the legs’ range of motion (RoM) of approximately $\pm135$ degrees allows the legs to push off in either direction, initiating the flipping action by applying a reaction force against the ground or the opposing segment. This bi-directional RoM is essential, as the orientation of each segment relative to the ground changes continuously during the cyclic flipping sequence. Electrical power (5 VDC) is supplied through a boost-up module that draws power from a single 18650 battery.

The robot’s movement is controlled remotely via a web-based panel. The panel provides a set of button elements, each triggering a fetch request to the onboard ESP32 microcontroller. The ESP32 processes these commands and relays instructions to the motor driver and the servo motors. This motor driver directly regulates the movement of the robot’s wheels, while servo commands control the flipping legs.

One body segment is equipped with a small set of wheels, which allows the robot to reposition itself (such as reorienting before a flip) and to locomote efficiently across flat terrain. This wheeled mode is contrasted with the flipping mode, which is activated to traverse obstacles, gaps, or rough terrain. This dual-mode capability provides a clear basis for comparing the relative effectiveness of rolling versus flipping under different terrain conditions.

Additionally, one segment carries a small onboard camera, which streams real-time video back to the user’s control interface. This camera assists the human operator during remote navigation, particularly when repositioning the robot prior to flips or when conducting visual inspection tasks. This manual, camera-assisted teleoperation highlights both the current simplicity of the control system and the potential for future development of autonomous or sensor-guided control strategies.

\begin{figure}
\begin{center}
    \includegraphics[width=8.8cm]{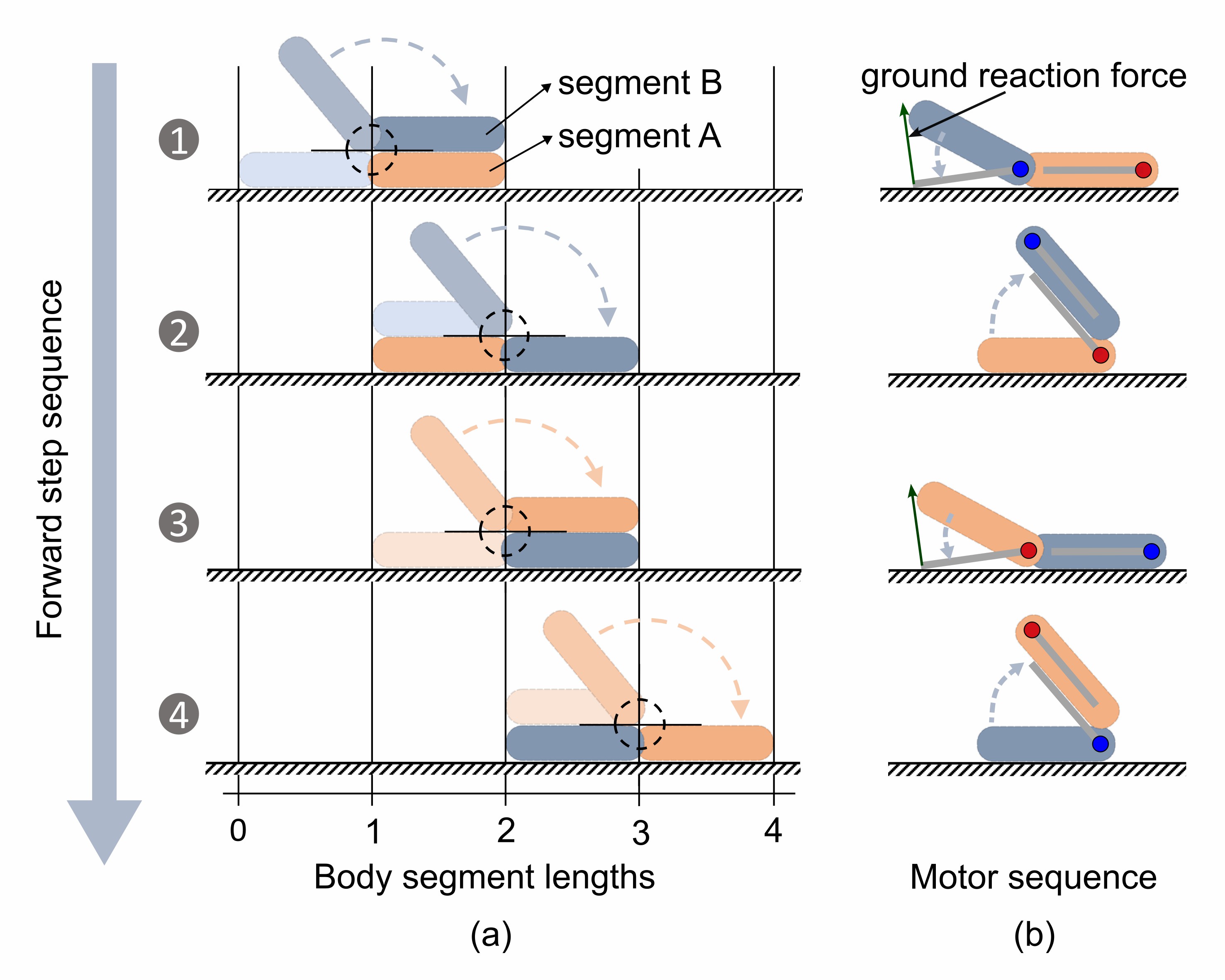}
\end{center}
\caption{A complete flip-walking cycle consists of four discrete steps shown in (a). Each step is driven by a rotational motor, which actuates a 1-DoF leg to push against either the ground surface or the opposing body segment of the robot. (b). The green arrow indicates the ground reaction force generated by the deployable flipper’s push-off action, which provides the necessary impulse to initiate each flip. This cyclic interaction between the flipper, the ground, and the robot body enables continuous locomotion across diverse terrains.}
\label{fig:sequence}
\end{figure}

\subsection{Forward locomotion sequence}
The style of forward locomotion of the robot is termed here 'flip-walking', as the robot moves forward by flipping over itself in discrete steps. A full flip-walking sequence is comprised of four steps, as shown in Fig.~\ref{fig:sequence}. In the first step of this sequence, the robot is initially configured with each body segment resting on a ground surface. The trailing segment (seg-B) is first to flip over the leading segment (seg-A) by extending its legs downward against the ground to cause a rotation about the top edge of contact between the two segments. The axis of rotation is depicted in the figure as a cross-hair icon. In this new configuration, seg-B is now resting on top of seg-A In the second step of flip-walking, seg-B flips again, but this time the legs of seg-B extend upward to cause seg-B to flip, and the rotation is moved to a new axis at the opposite side of seg-A compared to the previous step. At the termination of this step, seg-A is now positioned in the leading (front) position of the two blocks, which are now both resting on the ground again. From this configuration, as similar sequence of steps is performed as before, but now with seg-B remaining stationary, while seg-A flips over it twice (during two separate steps).  Again, the first flip is powered by the rotation of the leg of the active, flipping segment downward against the ground, while the second step occurs as a result of the upward rotation of the legs of the inactive, stationary segment. Also as in the first two steps, the center of rotation of these two flipping steps switches from the rear of the stationary segment (seg-B) to the front edge, as a result of the cable singularity mechanism.

\section{Analysis}
Here we derive a model which relates critical parameters to robot performance, allowing for the design and selection of components to match operational requirements, including obstacle clearance and minimum motor power. The model specifically focuses on the most common and stable climbing configurations observed during the flipping process.
\subsection{Analytical model}
\begin{figure}[t]
    \centering
    \includegraphics[width=8.8cm]{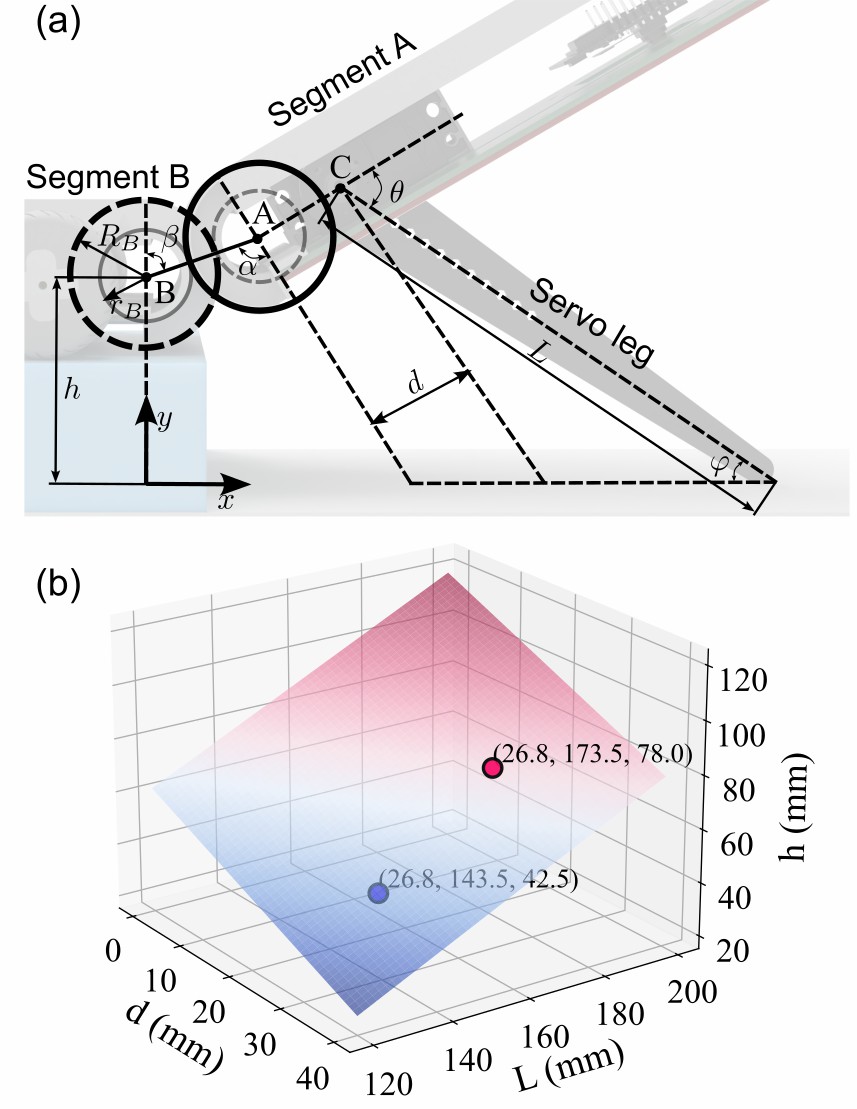}
    \caption{Analysis of FlipWalker’s joint: (a) geometric parameters of a FlipWalker and (b) the influence of $d$ and $L$ on the maximum climbing height $h$ during flipping. The red dot at coordinates (26.8, 173.5, 78.0) corresponds to Model~L (longer legs), while the blue dot at (26.8, 143.5, 42.5) represents Model~S (shorter legs).}
    \label{fig:analytical}
\end{figure}

We consider points $A$ and $B$ the interface center for segment A and segment B respectively, where joint rotation is achieved via an effective rolling contact point between two semicircular profiles so that the rotation does not occur about a fixed axis but along a moving instantaneous center of rotation which translates along the curved profile. For purposes of modeling and coordinate reference we annotate the geometric center of the rounded segment ends as the Interface Center of the corresponding segments. However, this does not imply that rotation occurs explicitly at this point. $C$ is the center of the rotary axes for the servo motor. The distance $d$ between $A$ and $C$ specifies the servo motor's installation position, and $L$ denotes the length of leg A (see Fig.~\ref{fig:analytical}(a)). The climbing height $h$ is defined as the distance between $B$ and the ground plane.

To describe the system’s geometry, we define the coordinate system with the origin at ground level directly below point $B$, the x-axis oriented horizontally forward, and the y-axis vertically upward. Accordingly, the coordinates of the key points are:

\begin{itemize}
    \item \textbf{Point $B$ (Segment B Interface Center):} $(x_B, y_B) = (0, h)$
    \item \textbf{Point $A$ (Segment A Interface Center):} 
    \[
    (x_A, y_A) = \bigl((r_b + R_b)\cos(\alpha-\frac{\pi}{2}),\,(r_a + R_b)\sin(\alpha-\frac{\pi}{2}) + h\bigr)
    \]
    \item \textbf{Point $C$ (Servo Motor Rotation Center):}
    \[
    (x_C, y_C) = \bigl(x_A + d \cos(2\alpha - \pi),\; y_A + d \sin(2\alpha - \pi)\bigr)
    \]
\end{itemize}

where $r_a = r_b = 12\,\text{mm}$ and $R_a = R_b = 20\,\text{mm}$ are the internal and external radii of two segments from the CAD model. Due to the constraints of cable~1 and cable~2, we have $\alpha + \beta = \pi$. From geometric considerations, the leg-contact angle is given by $\varphi = \pi + \theta - 2\alpha$.

Combining these relationships yields,
\begin{equation}
    \frac{y_C}{L} = \sin\varphi.
\label{eq:h_gov1}
\end{equation}
Substituting $y_C$ and $\varphi$ into \eqref{eq:h_gov1} gives,
\begin{equation}
    \frac{(32 \sin\alpha + h) + d \sin(2\alpha - \pi)}{L} = \sin(\pi + \theta - 2\alpha).
\label{eq:h_gov2}
\end{equation}
Solving \eqref{eq:h_gov2} for $h$ can be achieved by,
\begin{equation}
    h = \,L\,\sin(2\alpha - \theta)\;+\;32\,\cos\alpha\;+\;d\,\sin(2\alpha),
\label{eq:h_solution}
\end{equation}
which expresses $h$ in terms of $d$, $L$, $\alpha$, and $\theta$.

We impose that $\gamma \geq \tfrac{\pi}{2}$, where $\gamma$ is the angle between vector $\mathbf{AC}$ and the $y$-axis. By definition,
\begin{equation}
    \gamma = \arccos\!\biggl(\frac{\mathbf{AC}\cdot\mathbf{e_y}}{|\mathbf{AC}|\;|\mathbf{e_y}|}\biggr),
\end{equation}
with the $y$-axis unit vector $\mathbf{e_y} = (0,1)$. From the geometric configuration, we find $\gamma = 2\alpha - \pi$.

Given the servo motor’s range of motion (RoM) of 270°, the maximum allowable angle is $\theta_{\max} = \tfrac{\mathrm{RoM}}{2} = 135^\circ$.

Substituting $\theta_{\max}$ and $\gamma_c$ into \eqref{eq:h_solution} yields the 3D surface plot shown in Fig.~\ref{fig:analytical}(b). The horizontal axes represent the design parameters $d$ and $L$, while the vertical axis denotes the predicted climbing height $h$. The color gradient, ranging from blue to red, indicates increasing values of $h$. As $d$ decreases and $L$ increases, $h$ transitions from lower (blue) to higher (red), which aligns with the analytical prediction derived from equation \eqref{eq:h_solution}.

Based on the parameters listed in Table~\ref{table1}, we fabricated two robot prototypes with different leg lengths, denoted as Model~S (shorter legs) and Model~L (longer legs). Experimental trials were conducted to measure the actual climbing height, denoted as $h_{trial}$. The results demonstrate that Model~L achieved better climbing performance than predicted by the analytical model, whereas Model~S performed slightly worse than expected.

This outcome aligns with our expectations: the analytical model does not account for the kinetic energy contribution during flipping, so the actual climbing height is expected to slightly exceed the static prediction. The performance discrepancy observed in Model~S is likely attributed to the smaller radius at the leg tip, which caused the actual leg-contact angle $\varphi$ to be smaller than the analytical estimate. This issue was addressed in Model~L by increasing the leg tip radius, leading to improved agreement between the experimental and analytical results.

\begin{table}
\begin{center}
\begin{tabular}{ l l l }
 \hline
 Parameter \& Performance & Value (Model S/L)  & Unit \\ 
 \hline
 Maximum flipping speed &  0.2/0.2 & Body Length/s \\ 
 Maximum motor-driving speed &  1.10/0.95 & Body Length/s \\ 
 Dimensions for Model S &  $220\times75\times100$ & mm \\ 
 Dimensions for Model L &  $250\times75\times100$ & mm\\
 $d$  &  26.8/26.8 & mm\\
 $L$  &  143.5/173.5 & mm\\
 Weight &  708.43/777.78 & grams\\
 $h_{trial}$ &  42.5/78.0 & mm\\
 \hline
\end{tabular}
\end{center}
\caption{Actual parameter values from robot prototype Model S/L}
\label{table1}
\end{table}

\subsection{Dynamics}
The flipping dynamics of FlipWalker can be formulated using the Lagrange equation, which relates the kinetic and potential energies of the system to derive its equations of motion. As the axis of rotation between the two body segments changes with each flip, it is impractical to apply a direct actuation torque at this inter-segment joint. Instead, the flipping motion is initiated by a ground reaction force applied through the extensible leg, which pushes off against the stationary environment, either the ground surface or the opposing body segment, depending on the robot's configuration.

\begin{figure}[t]
\begin{center}
    \includegraphics[width=8.8cm]{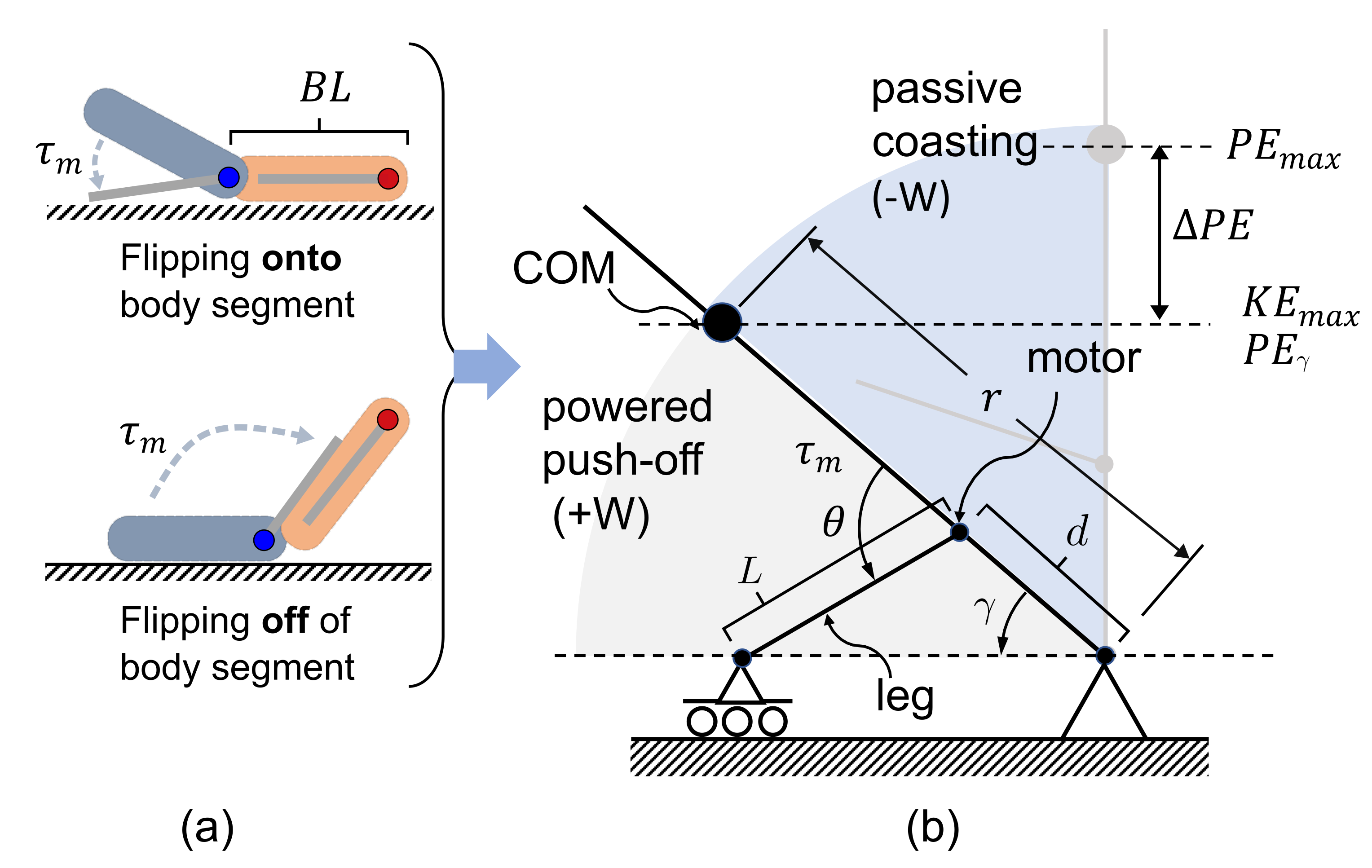}
\end{center}
\caption{Flipping occurs from two different configurations of the robot (a), but the mechanics and dynamics of actuation are the same in each case, shown in (b) to be dependent on physical design parameters which can be approximated by a simplified model.}
\label{fig:dynamics}
\end{figure}

The generalized coordinate $\gamma$ denotes the rotation angle of the active segment relative to the horizontal ground. The total kinetic energy $T$ of the system includes the rotational kinetic energy of the flipping segment:

\begin{equation}
    T = \frac{1}{2} I \dot{\gamma}^2
\end{equation}

where $I$ is the moment of inertia of the flipping body segment about the rotational axis.

The potential energy $V$ is the gravitational potential energy of the segment’s center of mass (COM), located at distance $r$ from the rotational axis:

\begin{equation}
    V = -mgr\sin\gamma
\end{equation}

The Lagrangian $\mathcal{L}$ is then:

\begin{equation}
    \mathcal{L} = T - V = \frac{1}{2}I\dot{\gamma}^2 + mgr\sin\gamma
\end{equation}

Applying the Euler-Lagrange equation:

\begin{equation}
    \frac{d}{dt}\left(\frac{\partial \mathcal{L}}{\partial \dot{\gamma}}\right) - \frac{\partial \mathcal{L}}{\partial \gamma} = \tau_{\mathrm{leg}}
\end{equation}

where $\tau_{\mathrm{leg}}$ is the external torque applied by the push-off force from the actuated leg. The leg applies this torque only during its active stroke, which spans a limited range of motion, $\theta_{max}$:

\begin{equation}
    \tau_{\mathrm{leg}} = 
    \begin{cases}
        \tau_m, & 0 \leq \theta \leq \theta_{max} \\
        0, & \theta > \theta_{max}
    \end{cases}
\end{equation}

Here, $\theta_{max}$ denotes the "cut-off" angle where the leg reaches its full extension and breaks contact with the ground. After this point, the segment relies entirely on its accumulated kinetic energy and gravitational potential to complete the flip.

The mechanical work done by the motor during the push-off phase is:

\begin{equation}
    W_m = \int_0^{\theta_{max}} \tau_m \, d\theta = \tau_m \theta_{max}
\end{equation}

To successfully complete a flip, the total energy at $\gamma$ must be sufficient to carry the body past vertical ($\gamma = 90^\circ$). This requires that:

\begin{equation}
    W_m \geq E_c = \Delta{PE}=PE_{max}-PE_{\gamma}
\end{equation}

The required relationship between motor torque, leg geometry, and climbing height then becomes:

\begin{equation}\label{eq:required_work}
    \tau_m \theta_{max} \geq mgr \left(1 - \sin\gamma\right)
\end{equation}

The geometric coupling between $\theta_{max}$ and $\gamma$ is derived from the relative placement of the leg and body segment, approximated by the 1-D rigid body configuration shown in Fig.~\ref{fig:dynamics}. This gives:

\begin{equation}
    d\sin\gamma = L\sin\left(\theta_{max} - \gamma\right)
\end{equation}

This system of equations describes how the leg motor’s work translates into flipping motion, thereby relating the key design parameters to the robot’s obstacle clearance capability: motor torque $\tau_m$, leg length $L$, mounting offset $d$, and actuation range $\theta_{\max}$. By quantifying the minimum torque required to propel a body segment beyond the vertical (i.e., to complete a dynamic, underactuated flip), this analytical framework provides a direct link between design decisions and performance. In practice, one could use the same relationships to determine, for a given motor torque, the minimum leg range of motion necessary to achieve a successful flip. As such, these equations enable informed design optimization for specific operational requirements, especially where stable climbing configurations are desired.

\section{Experiments}
\subsection{Terrain adaptability}
\begin{figure}[htbp]
\includegraphics[width=8.8cm]{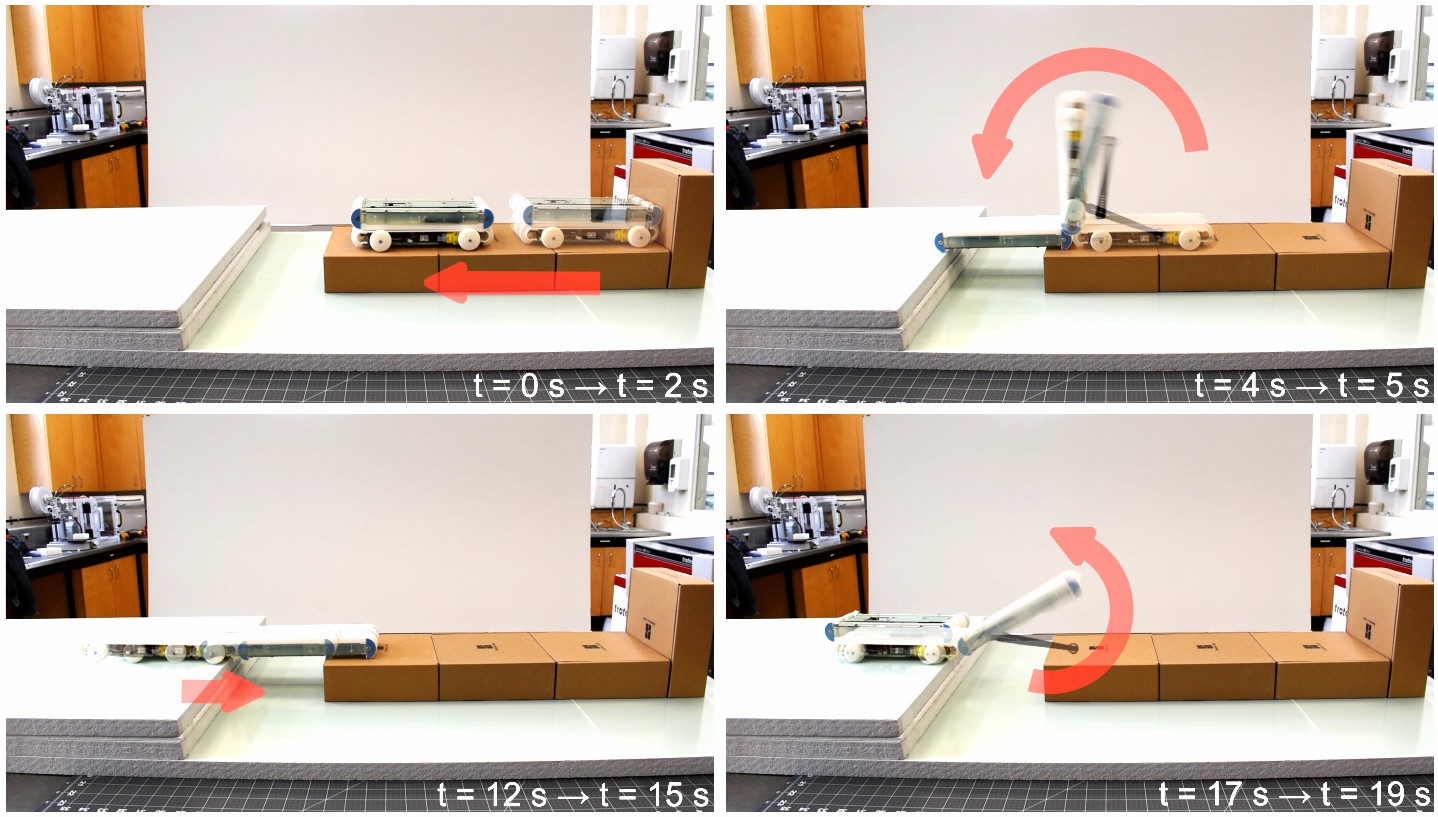}
\caption{Illustrating the robot bridging a gap between two surfaces using its deployable flipping mechanism. The red arrows indicate the direction of travel, and the curved arrow shows the flipping motion as the robot transitions from one platform to the other. }
\label{fig:gap}
\end{figure}

As shown in Fig.~\ref{fig:gap}, the FlipWalker robot employs its deployable flipping mechanism to traverse a 180 mm gap, wider than the 153.5 mm distance between its own wheels. At $t=0 s$, the robot is positioned on a raised platform serving as its starting point. Between $t=0 s$ and $t=2 s$, FlipWalker drives forward (red arrows) until it nears the edge. Then, from $t=4 s$ to $t=5 s$, it initiates a flipping maneuver, rotating its upper body to form a temporary “bridge”. After completing several flips, the robot repositions itself from $t=12 s$ to $t=15 s$ so that its servo leg can touch the original side of the gap. Finally, at $t=17 s$ to $t=19 s$, it performs a second flip to fully ascend onto the opposite surface. This sequence highlights FlipWalker’s ability to leverage repeated flipping motions and multi-segment flexibility to span substantial gaps without contacting the ground, which is an advantage not typically found in conventional wheeled or tracked robots.

\begin{figure}[htbp]
\includegraphics[width=8.8cm]{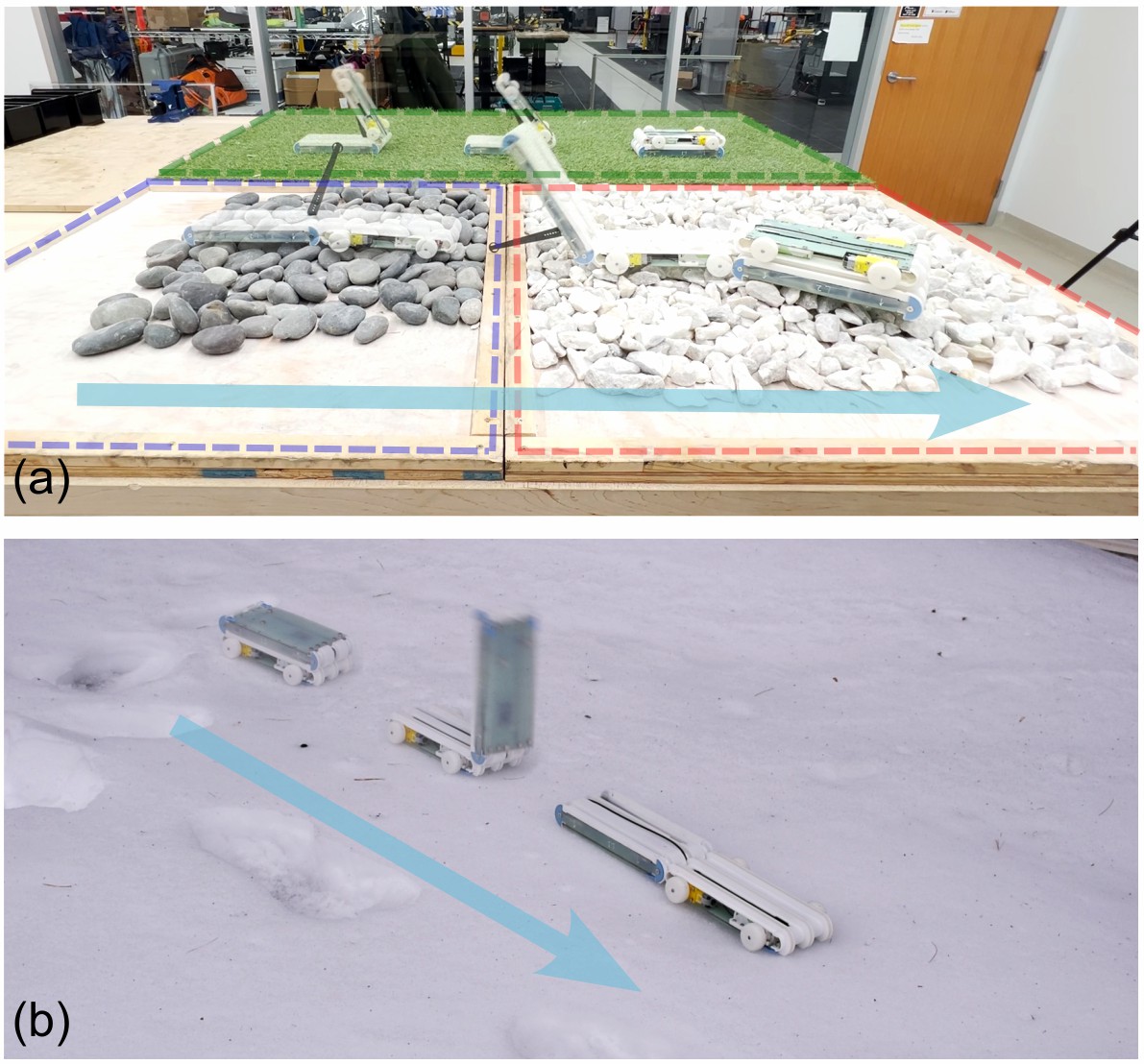}
\caption{The robot prototype was tested experimentally on four different types of terrain where wheeled systems often struggled. (a) Artificial grass, river rocks (purple frame), larger white marble stones (red frame), The arrows and colored outlines indicate the sequence of the transition of the robots between different surfaces. (b) Outdoor snowy ground. }
\label{fig:terrain}
\end{figure}

In addition to these gap-crossing demonstrations, the robot was also evaluated on a range of challenging terrain types where wheeled systems typically encounter difficulty. Specifically, the prototype was tested on four different surfaces: (1) artificial grass, (2) river rocks, (3) larger white marble stones, and (4) natural snow (outdoor, -4$^{\circ}$C), as shown in Fig.~\ref{fig:terrain}(a,b) and the Supplementary Video. 

For the natural snow experiments, FlipWalker was equipped with an easy-change magnetic flexible plate, which increased the effective ground contact area to nearly the size of the robot itself. This modification proved to be an effective strategy for improving mobility on soft, loose surfaces such as snow or sand, by distributing the contact pressure and reducing the risk of sinking.

\subsection{Multi-robot collaboration}
\begin{figure}[htbp]
\centering
\includegraphics[width=8.8cm]{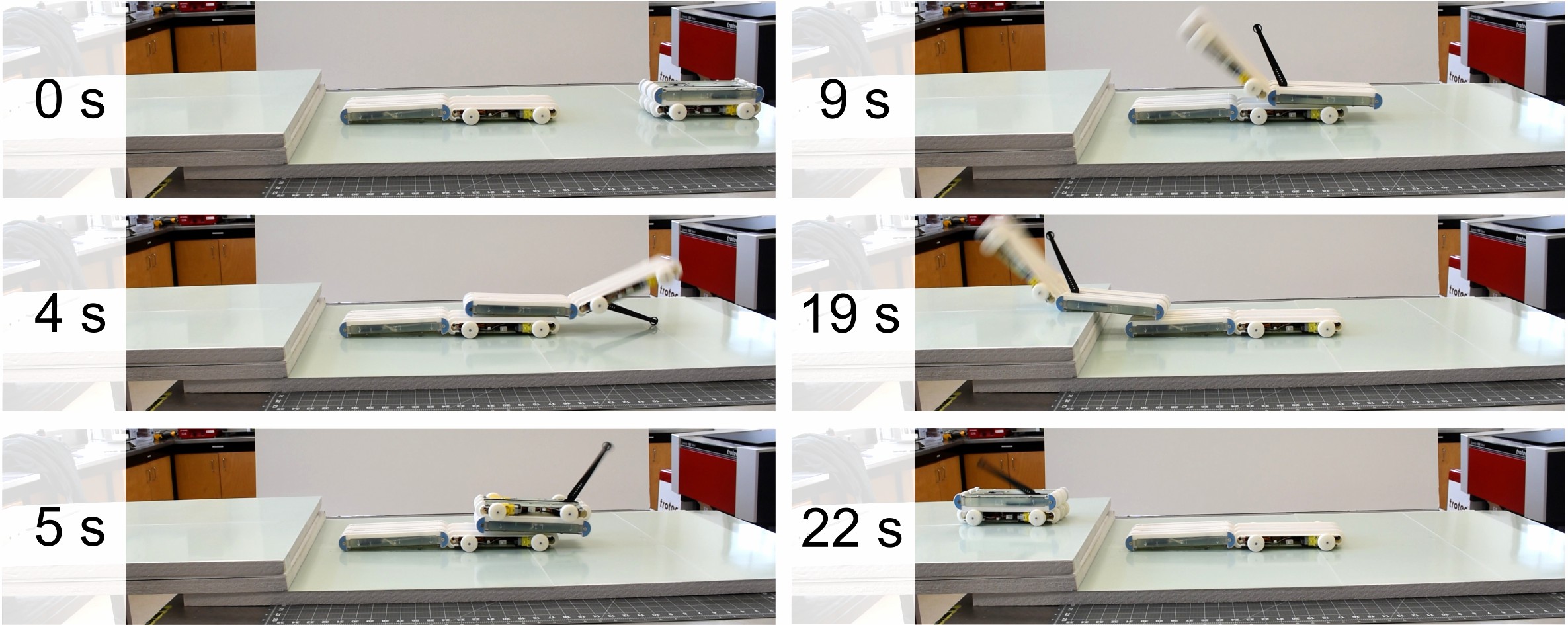}
\caption{Two Flipwalkers climbing the stage cooperatively.}
\label{fig:collab1}
\end{figure}

\begin{figure}[htbp]
\includegraphics[width=8.8cm]{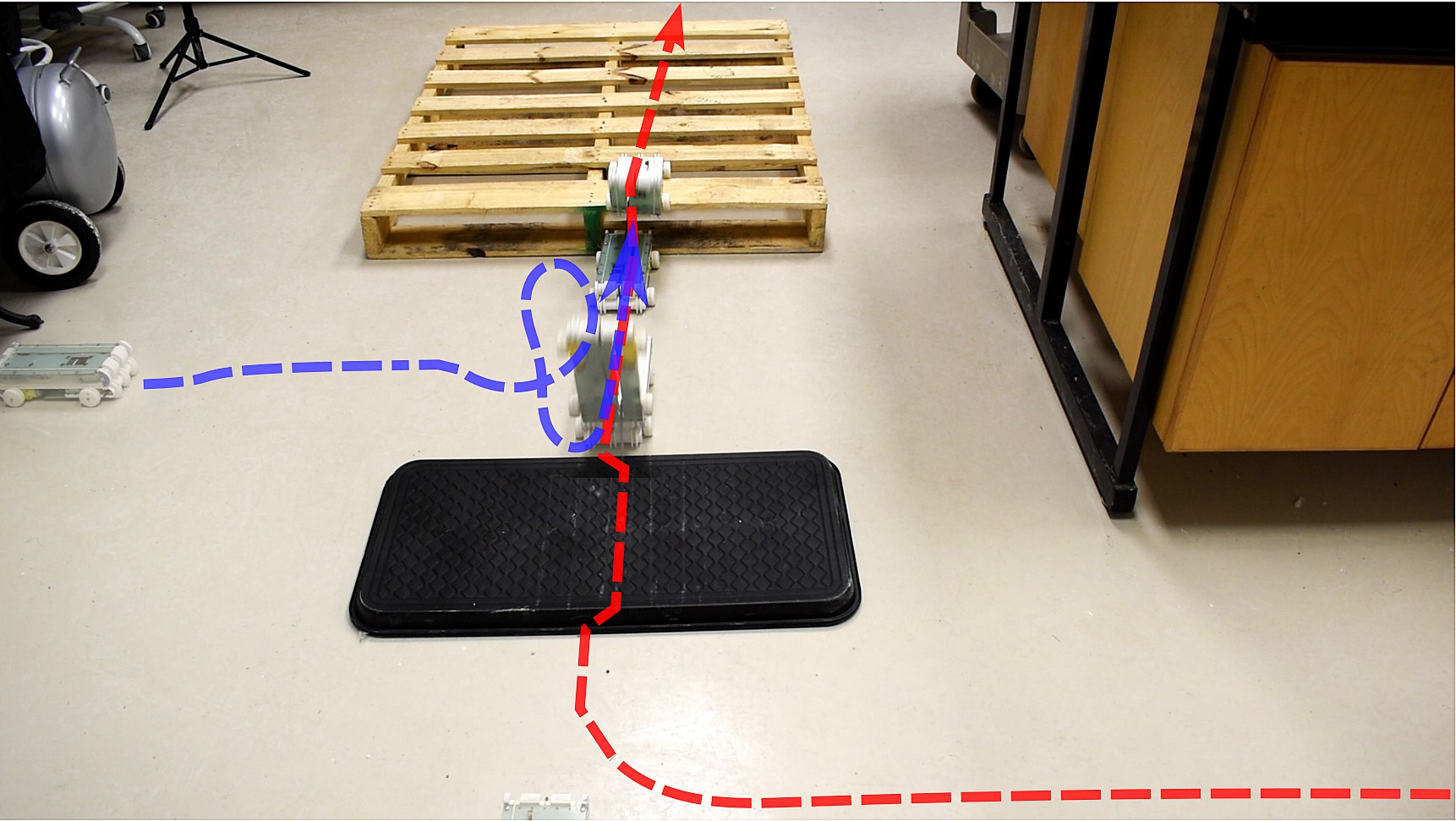}
\caption{Demonstrates the robot navigating over differently sized platforms and obstacles. Colored arrows trace the robot’s path, highlighting how two robots collaborate with each other to reach a higher surface, and realign their structure to overcome height difficulty.}
\label{fig:collab2}
\end{figure}

In addition to single-robot trials, we also investigated cooperative behaviors with multiple FlipWalker units. Fig.~\ref{fig:collab1} shows two FlipWalkers simultaneously climbing up an elevated stage with a climbing height of 80.0 mm. They coordinate their flipping actions to help one another gain the necessary leverage to overcome the height difference. As illustrated in Fig.~\ref{fig:collab2}, the climbing height of wood frame is 138.70 mm, and each robot’s trajectory (indicated by colored arrows) demonstrates how they strategically position themselves to share its height, allowing the robot to ascend to the stage. Through this collaborative approach, the FlipWalkers can tackle larger obstacles than would otherwise be possible with a single unit operating alone.

The second collaboration experiment, shown in Fig.~\ref{fig:collab2}, also evaluates FlipWalker’s collaborative capabilities under more complex, multi-obstacle scenarios. When confronted with steps and blockages of different sizes, the two FlipWalkers divide their tasks: one FlipWalker first arrives at the obstacle, adjusts its orientation and flipping configuration to serve as a ramp or anchor, and then the second FlipWalker uses this support to flip onto a higher step. Throughout this process, the robots serve two distinct purposes and roles as 'assistant' and 'climber' respectively, timing their flipping actions to accomplish what a single FlipWalker alone cannot achieve.

Notably, the wooden frame used in this test features open slots and gaps between its beams, which would pose significant challenges for conventional wheeled robots. These gaps interrupt continuous rolling contact, making traction unreliable. In contrast, FlipWalker’s flipping locomotion relies primarily on discrete push-off events and reconfiguration, allowing it to traverse such irregular structures with minimal reliance on surface continuity.

\subsection{Cost of transport (CoT)}

We used an INA226 module (Texas Instruments) to measure the total power consumption of our model L robot while traversing a distance of 10 body lengths (2.5\,m) on flat epoxy floor. The measured data was wirelessly transmitted to a PC for real-time analysis. The results show that in flipping mode, the robot achieves a COT of 16.98, whereas in wheel driving mode, the COT is reduced to 3.27. Additionally, the system's idle power consumption (when stationary) was measured to be 2.58\,W. The COT is calculated using the following equation:
\begin{equation}
\mathrm{CoT} = \frac{E}{m\,g\,d},
\label{eq:cot}
\end{equation}
where \(E\) is the total energy consumption in joules, \(m\) is the mass of the robot in kilograms, \(g = 9.81\,\mathrm{m/s^2}\) is the gravitational acceleration, and \(d\) is the distance traveled in meters. The following quantitative comparison helps highlight the relationships between platform size, design parameters, and energy efficiency, which are key considerations in field robotics research.

Table \ref{table2} presents a comparative overview of representative robots categorized by their total weight and corresponding Cost of Transport (COT). The robots are grouped into three classes based on their mass requirements and typical operational environments: (1) Field robots exceeding 10\,kg, (2) Portable field robots under 10\,kg, and (3) Pocket field robots below 1\,kg. Within each group, the table lists each platform’s measured or reported COT, often reflecting different locomotion modes (e.g., rolling vs.\ driving), together with the vehicle’s overall mass. Notably, ANYmal~\cite{hutter2016anymal} and Cheetah-Cub~\cite{sprowitz2013towards} appear in different categories despite both being legged systems, illustrating that size alone can shape operating constraints and performance metrics. For multi-modal platforms such as GOAT~\cite{polzin2025robotic}, the distinct COT values highlight the varying energy costs across locomotion methods. Similarly, the ART system~\cite{baines2022multi} reveals a broader COT range, underscoring the adaptability and complexity of multi-functional legged architectures. The M4 design~\cite{sihite2023multi} also spans a relatively broad COT range, suggesting the impact of usage scenarios or mechanical configurations. Finally, Flipwalker, a sub-kilogram pocket robot, demonstrates that miniaturized designs can still achieve mobility but may incur higher COT in certain gaits (e.g., flipping vs.\ driving).

\begin{table}[h]
\centering
\caption{Comparison of Robots: Cost of Transport (COT) and Weight}
\begin{tabular}{l c c}
\hline
\textbf{Robots} & \textbf{COT} & \textbf{Weight (kg)} \\ \hline
\multicolumn{3}{c}{Field robot ($>10\,\mathrm{kg}$)} \\ \hline
ANYmal\cite{hutter2016anymal}      & 1.2          & 30  \\ \hline
\multicolumn{3}{c}{Portable field robots ($<10\,\mathrm{kg}$)} \\ \hline
Cheetah-Cub \cite{sprowitz2013towards}   & 6.9          & 1.1 \\ 
M4\cite{sihite2023multi}          & 2.6 – 27.0            & 5.6 \\ 
GOAT\cite{polzin2025robotic}        & 0.1  (rolling) / 0.8  (driving)    & 1.6 \\ 
ART\cite{baines2022multi}         & 10.6 - 30.0            & 9   \\ \hline
\multicolumn{3}{c}{Pocket field robot ($<1\,\mathrm{kg}$)} \\ \hline
Flipwalker  & 16.98 (flipping) / 3.27 (driving) & 0.77 \\ \hline
\end{tabular}
\label{table2}
\end{table}

\section{Discussion and Conclusion}
Qualitatively, the experimental testing over different terrain types shown in this work demonstrated that the robot prototype specifically, and the novel flip-walking concept more generally, is capable of locomotion across diverse environmental conditions. The new form of locomotion presented offers new potential for developing simple, low-cost, yet effective mobile robots which can operate in a diverse range of terrain conditions. The simplicity of this design is amenable both to more robust robot platforms and to scalability at the individual robot level, and for larger collections of smaller, cost-effective robots (swarms). 

The current implementation relies on single-degree-of-freedom rotational legs to deliver the necessary push-off force for flipping. However, the overall structure and singularity-based joint mechanism derived from the Jacob’s Ladder toy can support alternative actuation schemes. For instance, solenoid actuators may deliver short duration, high impulse forces, while soft inflatable actuators could provide greater motion range or distribute forces more gradually. These approaches may also help improve energy efficiency, controllability, or terrain adaptability in future designs.

The current dynamic model does not include energy losses due to terrain, friction, or impact, which can significantly affect performance and will be addressed in future experimental modeling. Also, frequent reorientation from flipping complicates localization and control. To mitigate this, we plan to integrate inertial sensors to estimate body configuration and automate flip timing. Future work may explore sensing and coordination strategies for autonomous multi robot behaviors such as collaborative climbing, extending FlipWalker’s capabilities in real world scenarios.

\section*{Acknowledgment}
Funding was provided by the Natural Sciences and Engineering Research Council (NSERC) Discovery Grant RGPIN-2021-04049. Experimental testing was performed in part at the Ingenuity Labs Research Institute at Queen's University. The authors thank the following collaborators for their support and assistance in experimental testing trials: Zihuan Wu, Graziella Bedenik, Junhui Li, Hebe Iam and Timothy Byles-Ho.



\bibliographystyle{IEEEtran}
\bibliography{bibtex/ref}

\begin{thebibliography}{10}
\providecommand{\url}[1]{#1}
\csname url@samestyle\endcsname
\providecommand{\newblock}{\relax}
\providecommand{\bibinfo}[2]{#2}
\providecommand{\BIBentrySTDinterwordspacing}{\spaceskip=0pt\relax}
\providecommand{\BIBentryALTinterwordstretchfactor}{4}
\providecommand{\BIBentryALTinterwordspacing}{\spaceskip=\fontdimen2\font plus
\BIBentryALTinterwordstretchfactor\fontdimen3\font minus \fontdimen4\font\relax}
\providecommand{\BIBforeignlanguage}[2]{{%
\expandafter\ifx\csname l@#1\endcsname\relax
\typeout{** WARNING: IEEEtran.bst: No hyphenation pattern has been}%
\typeout{** loaded for the language `#1'. Using the pattern for}%
\typeout{** the default language instead.}%
\else
\language=\csname l@#1\endcsname
\fi
#2}}
\providecommand{\BIBdecl}{\relax}
\BIBdecl

\bibitem{chung_wheeled_2016}
W.~Chung and K.~Iagnemma, ``\BIBforeignlanguage{en}{Wheeled {Robots}},'' in \emph{\BIBforeignlanguage{en}{Springer {Handbook} of {Robotics}}}, ser. Springer {Handbooks}, B.~Siciliano and O.~Khatib, Eds.\hskip 1em plus 0.5em minus 0.4em\relax Cham: Springer International Publishing, 2016, pp. 575--594.

\bibitem{zheng2022mathbf}
C.~Zheng, S.~Sane, K.~Lee, V.~Kalyanram, and K.~Lee, ``$\alpha$-waltr: Adaptive wheel-and-leg transformable robot for versatile multiterrain locomotion,'' \emph{IEEE Transactions on Robotics}, vol.~39, no.~2, pp. 941--958, 2022.

\bibitem{lee_series_2016}
G.~Lee, H.~Kim, K.~Seo, J.~Kim, M.~Sitti, and T.~Seo, ``\BIBforeignlanguage{en}{Series of {Multilinked} {Caterpillar} {Track}-type {Climbing} {Robots}},'' \emph{\BIBforeignlanguage{en}{Journal of Field Robotics}}, vol.~33, no.~6, pp. 737--750, 2016.

\bibitem{kalra_wall_2006}
L.~P. Kalra, J.~Gu, and M.~Meng, ``A {Wall} {Climbing} {Robot} for {Oil} {Tank} {Inspection},'' in \emph{2006 {IEEE} {International} {Conference} on {Robotics} and {Biomimetics}}, Dec. 2006, pp. 1523--1528.

\bibitem{chen_soft_2023}
R.~Chen, X.~Tao, C.~Cao, P.~Jiang, J.~Luo, and Y.~Sun, ``A {Soft}, {Lightweight} {Flipping} {Robot} {With} {Versatile} {Motion} {Capabilities} for {Wall}-{Climbing} {Applications},'' \emph{IEEE Transactions on Robotics}, pp. 1--17, 2023.

\bibitem{liao_soft_2020}
B.~Liao, H.~Zang, M.~Chen, Y.~Wang, X.~Lang, N.~Zhu, Z.~Yang, and Y.~Yi, ``Soft {Rod}-{Climbing} {Robot} {Inspired} by {Winding} {Locomotion} of {Snake},'' \emph{Soft Robotics}, vol.~7, no.~4, pp. 500--511, Aug. 2020.

\bibitem{uno_hubrobo_2021}
K.~Uno, N.~Takada, T.~Okawara, K.~Haji, A.~Candalot, W.~F.~R. Ribeiro, K.~Nagaoka, and K.~Yoshida, ``{HubRobo}: {A} {Lightweight} {Multi}-{Limbed} {Climbing} {Robot} for {Exploration} in {Challenging} {Terrain},'' in \emph{2020 {IEEE}-{RAS} 20th {International} {Conference} on {Humanoid} {Robots} ({Humanoids})}, Jul. 2021, pp. 209--215, iSSN: 2164-0580.

\bibitem{guan_design_2022}
Y.~Guan, Z.~Zhuang, C.~Zhang, Z.~Tang, Z.~Zhang, and J.~S. Dai, ``\BIBforeignlanguage{en}{Design and {Motion} {Planning} of a {Metamorphic} {Flipping} {Robot}},'' \emph{\BIBforeignlanguage{en}{Actuators}}, vol.~11, no.~12, p. 344, Dec. 2022.

\bibitem{todd_walking_2013}
D.~J. Todd, \emph{\BIBforeignlanguage{en}{Walking {Machines}: {An} {Introduction} to {Legged} {Robots}}}.\hskip 1em plus 0.5em minus 0.4em\relax Springer Science \& Business Media, Mar. 2013.

\bibitem{torres-pardo_legged_2022}
A.~Torres-Pardo, D.~Pinto-Fernández, M.~Garabini, F.~Angelini, D.~Rodriguez-Cianca, S.~Massardi, J.~Tornero, J.~C. Moreno, and D.~Torricelli, ``\BIBforeignlanguage{en}{Legged locomotion over irregular terrains: state of the art of human and robot performance},'' \emph{\BIBforeignlanguage{en}{Bioinspiration \& Biomimetics}}, vol.~17, no.~6, p. 061002, Oct. 2022.

\bibitem{fong_mechanical_2005}
M.-f. Fong, ``\BIBforeignlanguage{eng}{Mechanical design of a simple bipedal robot},'' Thesis, Massachusetts Institute of Technology, 2005.

\bibitem{paluska_design_nodate}
D.~J. Paluska, ``\BIBforeignlanguage{en}{Design of a {Humanoid} {Biped} for {Walking} {Research}}.''

\bibitem{guizzo_by_2019}
E.~Guizzo, ``By leaps and bounds: {An} exclusive look at how {Boston} dynamics is redefining robot agility,'' \emph{IEEE Spectrum}, vol.~56, no.~12, pp. 34--39, Dec. 2019.

\bibitem{yim_hopping_nodate}
J.~Yim, ``\BIBforeignlanguage{English}{Hopping {Control} and {Estimation} for a {High}-{Performance} {Monopedal} {Robot}, {Salto}-{1P}},'' Ph.{D}., University of California, Berkeley, United States -- California, iSBN: 9798678171528.

\bibitem{fiorini_development_2003}
P.~Fiorini and J.~Burdick, ``\BIBforeignlanguage{en}{The {Development} of {Hopping} {Capabilities} for {Small} {Robots}},'' \emph{\BIBforeignlanguage{en}{Autonomous Robots}}, vol.~14, no.~2, pp. 239--254, Mar. 2003.

\bibitem{hyon_development_2002}
S.~Hyon and T.~Mita, ``Development of a biologically inspired hopping robot-"{Kenken}",'' in \emph{Proceedings 2002 {IEEE} {International} {Conference} on {Robotics} and {Automation} ({Cat}. {No}.{02CH37292})}, vol.~4, May 2002, pp. 3984--3991 vol.4.

\bibitem{brown_bow_1998}
B.~Brown and G.~Zeglin, ``The bow leg hopping robot,'' in \emph{Proceedings. 1998 {IEEE} {International} {Conference} on {Robotics} and {Automation} ({Cat}. {No}.{98CH36146})}, vol.~1, May 1998, pp. 781--786 vol.1, iSSN: 1050-4729.

\bibitem{yoshimitsu_micro-hopping_2003}
T.~Yoshimitsu, T.~Kubota, I.~Nakatani, T.~Adachi, and H.~Saito, ``Micro-hopping robot for asteroid exploration,'' \emph{Acta Astronautica}, vol.~52, no.~2, pp. 441--446, Jan. 2003.

\bibitem{armour_rolling_2006}
R.~H. Armour and J.~F.~V. Vincent, ``\BIBforeignlanguage{en}{Rolling in nature and robotics: {A} review},'' \emph{\BIBforeignlanguage{en}{Journal of Bionic Engineering}}, vol.~3, no.~4, pp. 195--208, Dec. 2006.

\bibitem{liang2020freebot}
G.~Liang, H.~Luo, M.~Li, H.~Qian, and T.~L. Lam, ``Freebot: A freeform modular self-reconfigurable robot with arbitrary connection point-design and implementation,'' in \emph{2020 IEEE/RSJ International Conference on Intelligent Robots and Systems (IROS)}.\hskip 1em plus 0.5em minus 0.4em\relax IEEE, 2020, pp. 6506--6513.

\bibitem{gim2024ringbot}
K.~G. Gim and J.~Kim, ``Ringbot: Monocycle robot with legs,'' \emph{IEEE Transactions on Robotics}, vol.~40, pp. 1890--1905, 2024.

\bibitem{seo2013flipbot}
B.~Seo, H.~Kim, M.~Kim, K.~Jeong, and T.~Seo, ``Flipbot: a new field robotic platform for fast stair climbing,'' \emph{International Journal of Precision Engineering and Manufacturing}, vol.~14, no.~11, pp. 1909--1914, 2013.

\bibitem{maslin_underwater_2021}
M.~Maslin, S.~Louis, K.~Godary~Dejean, L.~Lapierre, S.~Villéger, and T.~Claverie, ``\BIBforeignlanguage{en}{Underwater robots provide similar fish biodiversity assessments as divers on coral reefs},'' \emph{\BIBforeignlanguage{en}{Remote Sensing in Ecology and Conservation}}, vol.~7, no.~4, pp. 567--578, 2021.

\bibitem{aucone_drone_assisted_2023}
E.~Aucone, S.~Kirchgeorg, A.~Valentini, L.~Pellissier, K.~Deiner, and S.~Mintchev, ``Drone-assisted collection of environmental {DNA} from tree branches for biodiversity monitoring,'' \emph{Science Robotics}, vol.~8, no.~74, p. eadd5762, Jan. 2023.

\bibitem{angelini_robotic_2023}
F.~Angelini, P.~Angelini, C.~Angiolini, S.~Bagella, F.~Bonomo, M.~Caccianiga, C.~D. Santina, D.~Gigante, M.~Hutter, T.~Nanayakkara, P.~Remagnino, D.~Torricelli, and M.~Garabini, ``Robotic {Monitoring} of {Habitats}: {The} {Natural} {Intelligence} {Approach},'' \emph{IEEE Access}, vol.~11, pp. 72\,575--72\,591, 2023.

\bibitem{bogue_disaster_2019}
R.~Bogue, ``Disaster relief, and search and rescue robots: the way forward,'' \emph{Industrial Robot: the international journal of robotics research and application}, vol.~46, no.~2, pp. 181--187, Jan. 2019.

\bibitem{nagatani_emergency_2013}
K.~Nagatani, S.~Kiribayashi, Y.~Okada, K.~Otake, K.~Yoshida, S.~Tadokoro, T.~Nishimura, T.~Yoshida, E.~Koyanagi, M.~Fukushima, and S.~Kawatsuma, ``\BIBforeignlanguage{en}{Emergency response to the nuclear accident at the {Fukushima} {Daiichi} {Nuclear} {Power} {Plants} using mobile rescue robots},'' \emph{\BIBforeignlanguage{en}{Journal of Field Robotics}}, vol.~30, no.~1, pp. 44--63, 2013.

\bibitem{yang_grasping_2021}
Y.~Yang, K.~Vella, and D.~P. Holmes, ``Grasping with kirigami shells,'' \emph{Science Robotics}, vol.~6, no.~54, p. eabd6426, May 2021.

\bibitem{rus_spotlight_2018}
D.~Rus and C.~Sung, ``Spotlight on origami robots,'' \emph{Science Robotics}, vol.~3, no.~15, p. eaat0938, Feb. 2018.

\bibitem{robertson2021soft}
M.~A. Robertson, O.~C. Kara, and J.~Paik, ``Soft pneumatic actuator-driven origami-inspired modular robotic “pneumagami”,'' \emph{The International Journal of Robotics Research}, vol.~40, no.~1, pp. 72--85, 2021.

\bibitem{li2023origami}
D.~Li, D.~Fan, R.~Zhu, Q.~Lei, Y.~Liao, X.~Yang, Y.~Pan, Z.~Wang, Y.~Wu, S.~Liu \emph{et~al.}, ``Origami-inspired soft twisting actuator,'' \emph{Soft Robotics}, vol.~10, no.~2, pp. 395--409, 2023.

\bibitem{nelson_curved-folding-inspired_2016}
T.~G. Nelson, R.~J. Lang, S.~P. Magleby, and L.~L. Howell, ``\BIBforeignlanguage{en}{Curved-folding-inspired deployable compliant rolling-contact element ({D}-{CORE})},'' \emph{\BIBforeignlanguage{en}{Mechanism and Machine Theory}}, vol.~96, pp. 225--238, Feb. 2016.

\bibitem{halverson2010tension}
P.~A. Halverson, L.~L. Howell, and S.~P. Magleby, ``Tension-based multi-stable compliant rolling-contact elements,'' \emph{Mechanism and machine theory}, vol.~45, no.~2, pp. 147--156, 2010.

\bibitem{noauthor_scientific_1889}
\BIBentryALTinterwordspacing
\emph{\BIBforeignlanguage{eng}{Scientific {American} {Volume} 61 {Number} 15}}, Oct. 1889. [Online]. Available: \url{http://archive.org/details/scientific-american-1889-10-12}
\BIBentrySTDinterwordspacing

\bibitem{noauthor_physics_nodate}
\BIBentryALTinterwordspacing
``Physics {Toys}, {Tricks} and {Teasers}.'' [Online]. Available: \url{https://www.lockhaven.edu/~dsimanek/TTT-rings/rings.htm}
\BIBentrySTDinterwordspacing

\bibitem{immel_jacobs_2019}
A.~Immel, ``\BIBforeignlanguage{en-US}{The {Jacob}’s {Ladder} {Toy} and {Its} {Mysterious} {History}},'' Jan. 2019.

\bibitem{noauthor_david_nodate}
``David {Mitchell}'s {Origami} {Heaven} - {History} - {The} {Chinese} {Wallet} / {Jacob}'s {Ladder}.''

\bibitem{hirth_luca_2015}
T.~W. N. d.~S. Hirth, ``\BIBforeignlanguage{eng}{Luca {Pacioli} and his 1500 book de {Viribus} {Quantitatis}},'' 2015.

\bibitem{geng_dynamics_2005}
T.~Geng, ``Dynamics and trajectory planning of a planar flipping robot,'' \emph{Mechanics Research Communications}, vol.~32, no.~6, pp. 636--644, Nov. 2005.

\bibitem{hutter2016anymal}
M.~Hutter, C.~Gehring, D.~Jud, A.~Lauber, C.~D. Bellicoso, V.~Tsounis, J.~Hwangbo, K.~Bodie, P.~Fankhauser, M.~Bloesch \emph{et~al.}, ``Anymal-a highly mobile and dynamic quadrupedal robot,'' in \emph{2016 IEEE/RSJ international conference on intelligent robots and systems (IROS)}.\hskip 1em plus 0.5em minus 0.4em\relax IEEE, 2016, pp. 38--44.

\bibitem{sprowitz2013towards}
A.~Spr{\"o}witz, A.~Tuleu, M.~Vespignani, M.~Ajallooeian, E.~Badri, and A.~J. Ijspeert, ``Towards dynamic trot gait locomotion: Design, control, and experiments with cheetah-cub, a compliant quadruped robot,'' \emph{The International Journal of Robotics Research}, vol.~32, no.~8, pp. 932--950, 2013.

\bibitem{polzin2025robotic}
M.~Polzin, Q.~Guan, and J.~Hughes, ``Robotic locomotion through active and passive morphological adaptation in extreme outdoor environments,'' \emph{Science Robotics}, vol.~10, no.~99, p. eadp6419, 2025.

\bibitem{baines2022multi}
R.~Baines, S.~K. Patiballa, J.~Booth, L.~Ramirez, T.~Sipple, A.~Garcia, F.~Fish, and R.~Kramer-Bottiglio, ``Multi-environment robotic transitions through adaptive morphogenesis,'' \emph{Nature}, vol. 610, no. 7931, pp. 283--289, 2022.

\bibitem{sihite2023multi}
E.~Sihite, A.~Kalantari, R.~Nemovi, A.~Ramezani, and M.~Gharib, ``Multi-modal mobility morphobot (m4) with appendage repurposing for locomotion plasticity enhancement,'' \emph{Nature communications}, vol.~14, no.~1, p. 3323, 2023.

\end{thebibliography}
%

\end{document}